\documentclass[sigconf]{acmart}
\AtBeginDocument{%
  }

\copyrightyear{2026}
\acmYear{2026}
\setcopyright{cc}
\setcctype{by}
\acmConference[WebSci '26]{18th ACM Web Science Conference}{May 26--29, 2026}{Braunschweig, Germany}
\acmBooktitle{18th ACM Web Science Conference (WebSci '26), May 26--29, 2026, Braunschweig, Germany}
\acmDOI{10.1145/3795766.3799754}
\acmISBN{979-8-4007-2504-3/2026/05}

\usepackage{subcaption}

\newcommand{\model}{\textsc{PerspectiveGap}}
\newcommand{\paligned}{\texttt{bureaucrat}}
\newcommand{\critical}{\texttt{civilian}}
\newcommand{\context}{\texttt{context}}
\newcommand{\fastcoref}{\texttt{fastcoref}}
\newcommand{\flantfive}{\texttt{flan-t5}}
\newcommand{\entities}{\mathcal{E}}
\newcommand{\fcorefentities}{\hat{\mathcal{E}}}

\begin{document}

\title{Quantifying Media Representation Dynamics Across 25 Years of News Reporting on Policing-related Deaths}

\author{Farhan Samir}
\email{f.samir@utoronto.ca}
\affiliation{%
  \institution{University of Toronto}
  \city{Toronto}
  \country{Canada}
}

\author{Jappun Dhillon}
\affiliation{%
  \institution{University of British Columbia}
  \city{Vancouver}
  \country{Canada}}

\author{Meghna Ravikumar}
\affiliation{%
  \institution{University of Toronto}
  \city{Toronto}
  \country{Canada}
}

\author{Syed Ishtiaque Ahmed}
\affiliation{%
 \institution{University of Toronto}
 \city{Toronto}
 \country{Canada}}

\author{Vered Shwartz}
\affiliation{%
  \institution{University of British Columbia}
  \city{Vancouver}
  \country{Canada}}

\renewcommand{\shortauthors}{Trovato et al.}

\begin{abstract}
We perform the largest known computational analysis of Canadian news narratives about police-involved deaths, spanning 4,000 articles from the last quarter-century. We develop a novel computational model, \model{}, grounded in prior grounded in prior sociological work on media representation of policing. We find that reporting on police-involved deaths on average features perspectives from state bureaucrats at a rate nearly three times as much as perspectives from other members of the public, including relatives, community members, eyewitnesses, lawyers representing the family, or civil liberties groups. A considerable fraction of articles contain no points of view from civilian actors, though civilian representation has increased in recent years. 
Qualitatively, we find that state bureaucrats' accounts of these deaths tend to be clinical and procedural, while civilian discourse carries considerably more emotional valence. The \textsc{PerspectiveGap} framework developed here can be contextualized to other jurisdictions, offering a scalable approach for analyzing how media systems construct narratives around policing and accountability.\footnote{Project repository: \tiny{\url{https://github.com/smfsamir/perspective-gap}}} 
\end{abstract}




\maketitle

\section{Introduction}
\begin{figure}
    \centering
    \includegraphics[width=\linewidth]{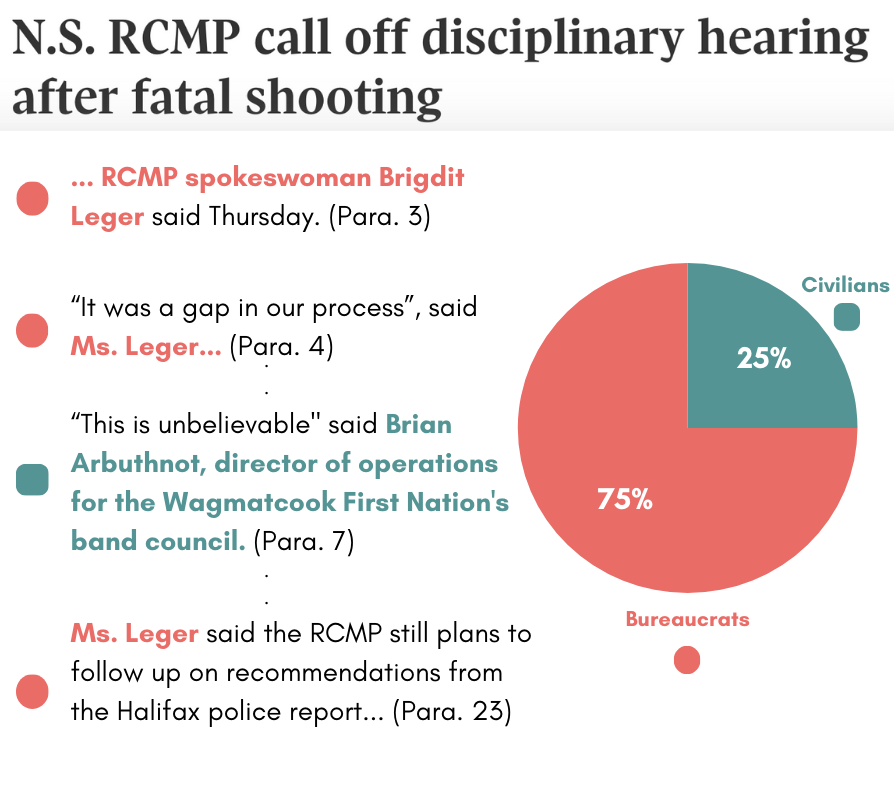}
    \caption{We measure the volume of perspectives from state bureaucrats in reporting on police-involved deaths, in comparison to civilian-based perspectives. This particular example is from \textit{The Globe and Mail} \citep{tutton2010rcmp}.}
    \Description{An annotated example of a Globe and Mail article titled "N.S. RCMP call off disciplinary hearing after fatal shooting," illustrating the labeling methodology. Selected sentences are highlighted in red (bureaucrat) or teal (civilian), with paragraph numbers noted. A pie chart shows the article is 75 percent bureaucrat and 25 percent civilian in perspective.}
    \label{fig:globe-and-mail-parse}
\end{figure}
News media organizations play an important role in the process of public opinion formation \citep{dewey_public_2012,lippmann_public_2017,rosen1999journalists,entman1993freezing,hall2017policing}. How exactly news organizations should go about informing the public has been widely debated.  \citet{lippmann_public_2017} famously argued for a more \emph{technocratic} form of press reporting, wherein institutional actors are elicited for their expert opinions, that are then relayed through the press. Conversely, \citet{rosen1999journalists} and \citet{dewey_public_2012}, among many others, critiqued at great lengths this format of newsmaking that treats the public as passive consumers of information. Instead, these scholars argued for a more \emph{co-productive} process \citep{ostrom1996crossing}  that  engages the public in conversation with institutional actors. 

In this work we look at how news organizations have fared in balancing between technocratic and more co-productive forms of news production. We consider a context where institutional actors are known to differ considerably from civilians \citep{sutton_thin_2021}: deaths associated with policing, whether during police custody or a deadly force incident.  
Specifically, we ask: have news organizations taken a more technocratic view in articulating such incidents \citep{lippmann_public_2017}, appealing to the views of policing bureaucrats to interpret the events? Or have they included civilian testimony  in their reporting? How these news media narratives are constructed is important, as most people do not have regular confrontational interactions with police, and thus news contributes a major way in which public opinion is formed  surrounding such interactions \citep{karakatsanis2025copaganda,chan2014racialization}. 

Answering these questions requires us to understand the various organizations that are involved in making up the institution of criminal punishment system, as bureaucrats from these organizations are often consulted in reporting on police-involved deaths. Prior work has largely focused on the media representation of police officers  \citep{crowl_measuring_2025,ziems_protect_2021,arora2025multi}. However, decades of punishment sociology research informs us that police officers are only one part of a much broader coalition of organizations that make up the criminal punishment bureaucracy, including oversight agencies, police unions, and elected officials, among others. Rather than the simplistic bifurcation of police and victims, we consider a broader superset of \textit{bureaucrats} and \textit{civilians} (Section~\ref{sec:entities}), measuring the degree that their points of view are represented in news narratives (Figure~\ref{fig:globe-and-mail-parse}).

We present a computational framework for accurately measuring this perspective weighting, grounded in contextualized punishment sociology research (Sections~\ref{sec:background} and~\ref{sec:entities}). We develop the \model{} model, a computational pipeline relying only on language models that can run on consumer-grade laptops, yet can accurately identify points of view (from bureaucrats and civillians) across long chains of passages. Despite training only on a modest amount of carefully annotated data, we find that our model performs as well as a prompt-tuned GPT-4o (Section~\ref{sec:model}).

We then apply \model{} at a larger scale to over $4,000$ articles from Canadian news media. We find that news reporting on police-involved deaths over the last 25 years has been highly technocratic, with bureaucrat testimonies represented at about three times the rate of civilian accounts. From 2020-2023 however, we observed a dynamic increase in coverage of civilian accounts. We also observe significant variation across outlets in the rates of civillian-centric coverage. 



\section{Related Work} \label{sec:background}
Our work is about measuring epistemic authority \citep{foucault2013archaeology}: how often do bureaucrats involved in instituting state punishment get represented in reporting on police-involved deaths? More specifically, we aim to measure the volume of points of view from such bureaucrats reported in these news stories, relative to the volume of civilian perspectives. Naturally, actors whose perspectives are represented more often have greater epistemic authority in setting the bounds of reasonable discourse \citep{karakatsanis2025copaganda}. 

Prior computational works have largely been interested in questions of framing \citep{entman1993freezing}, rather than representational volume. \citet{arora2025multi} use the generic media frames corpus \citep{card2015media}, to label images and text in a large news corpus. They find that news stories with an image that is evocative of crime also quote police officers frequently. \citet{crowl_measuring_2025} study how readers across the United States   perceive sentences in local news stories about policing, namely whether they understand the sentence to portray the police in a positive or negative light. 

\citet{ziems_protect_2021} look at a number of different frames on news reports of police-involved deaths in the United States, related to whether the victim in the case was for example reported as fleeing or having a record. They also measured the proportion of official sources and unofficial ones who are quoted in the news story. This is similar to our aim. However, they use a rudimentary definition of what constitutes official sources, only looking at police officers, and treating other entities, such as media relations officers (Figure~\ref{fig:globe-and-mail-parse}) as unofficial sources. This approach simultaneously significantly underestimates the volume of perspectives that can be attributed to bureaucrats, and overestimates the perspectives attributable to unofficial (or civillian) sources. There is thus a tendency among prior computational works to focus only on a proper subset of institutional actors, specifically police officers; see also \citet[][inter alia]{rho2023escalated,field2023developing} for other examples.  

\citet{sutton_thin_2021} comes closer to addressing the question of representational volume, using a definition of ``official sources'' that includes not only police officers but also media relations officers and police oversight agency officers. They then counted the number of articles that contain the point of view of an official source or an unofficial one, for a corpus with 1,364 Canadian news articles. However, their approach does not distinguish between an article that quotes a spokesperson once, in comparison to an article that quotes them several times. We aim to precisely measure variation at the level of individual articles (Figure~\ref{fig:globe-and-mail-parse}).
\section{Entities in Canadian News Articles} \label{sec:entities}
In order to measure the degree of technocratic reporting in newsreporitng on deadly force incidents, we need to first characterize the relevant experts and institutions. We draw on \citeauthor{karakatsanis2025copaganda}'s (\citeyear{karakatsanis2025copaganda}) notion of the \textit{punishment bureaucracy}, a concept denoting the coalition of public institutions that are directly related to the governance of criminal activity and accountability. The make up of this coalition will vary across geopolitical contexts. We focus on the make up of the punishment bureaucracy and their representation in news stories on police-involved deaths in the Canadian context, drawing on prior sociological research on the relationship between the punishment bureaucracy and the press. 


\subsection{Official Sources} \label{sec:punishment-bureaucrats}

\paragraph{Police departments.} Local departments are known historically to have had close relationships with newsreporters \citep[Chapter 4,][]{chan2014racialization}. It thus almost goes without saying that department officials will be present in large quantities in these articles \citep{sutton_thin_2021}. For example, 

\begin{quote}
``\textit{Vancouver police chief Jim Chu issued a statement expressing regret over Hubbard's death, but defending the officers' actions.}'' \citep{cbc2009policeclear}

\end{quote}

Indeed, it is widely known that departments typically have public relations officers who coordinate statements and correspond directly with news reporters after deadly force incident \citep{walby_examining_2022}. In an analysis over 85 press releases from police media units throughout Canadian jurisdictions, \citet{walby_examining_2022} find that these statements serve to manage the police department's reputational risk, employing a number of rhetorical strategies including euphemism, passivization, and providing limited information outside of administrative procedures.  

When the deadly force incident brings a considerable degree of scrutiny, or inquest proceedings reach an advanced stage (charges being laid), police union officials representing officers also provide perspectives that are reported by news organizations. Perspectives that typically seek to defend and uphold the epistemic legitimacy of the bureaucracy in serving the public interest \citep{walby_examining_2022}:

\begin{quote}
``\textit{The union representing York Regional Police believes the Crown has gone too far in its prosecution of Romano.}'' \citep{grimaldi2016crown}
\end{quote}
Unions are also involved in hiring defense lawyers when officers are charged with criminal offences. These lawyers perspectives are also reported in news articles.

\paragraph{Oversight agencies.} 
All Canadian provinces have an ``Independent critical investigation agency'' \citep[Table 1 in][]{walby_examining_2022}. These agencies conduct investigations into deadly force incidents, aiming to determine whether deadly force was legally permissible according to the criminal code, and are thus frequently mentioned in news reports \citep{sutton_thin_2021}.
These bureaucracies were created to provide accountability mechanisms \citep{owusu-bempah_black_2014}. 
In practice however, these oversight boards have been criticized for having a pro-police bias; in the the Special Investigations Unit (SIU), a ``majority of the investigators are White men, and over two-thirds of investigators have a police background'' \citep{laming_police_2022}. Analogous allegations of pro-police bias have been made in other jurisdictions \citep{maynard2025policing}. Nationwide, the CBC found in a study of 461 fatalities over an 18-year time period that ``only 2 resulted in a criminal conviction'' \citep{whitehead2023deadly}.

Many news reports citing these oversight agencies contain perspectives that foreground the officer's subjective experience: 
\begin{quote}
   \textit{``The officers honestly believed they were looking at an actual gun in the Complainant's possession. Though mistaken, their misapprehension was a reasonable one.''} \citep{nielsen2025ontario}
\end{quote}

\noindent This is in line with \citeauthor{sutton_thin_2021}'s findings that claims made by ``official'' sources reported in are often supportive, neutral, but rarely critical.

\paragraph{Coroners.} Coroners are often involved in the oversight process, investigating cause of death, often decreed by statute \citep{whitehead2023deadly}. While ostensibly clinical and innocuous, identifying causes of death is politically fraught. Consider the statement of a defence lawyer pertaining to the death of Abdirahman Abdi: 

\begin{quote}
   ``\textit{This is not a beating that caused the death of Mr. Abdi. Mr. Abdi died of a heart attack.}'' \citep{cbc2017abdi}
\end{quote}

In a study on over 400 coroners reports related to deadly force incidents across all Canadian provinces, \citet{whitehead2023deadly} found that coroners' reports regularly employ rhetorical strategies that shift attention away from police culpability and towards the victim's behavior or mental state, constructing the use of force as an inevitable response rather than a discretionary act.

\paragraph{Prosecutors.} Another noteworthy authority involved in the punishment bureaucracy are prosecutors, or \textit{Crown attorneys} in the Canadian context. \citet{puddister2021police} investigate patterns in prosecutorial decisions of crown attorneys in cases where the aforementioned oversight agencies suggest that the officer under investigation be charged. They note, drawing on research situated in the US, that prosecutors are hesitant to charge officers. They go on to find in their study on 159 investigations over 15 years (2005-2020) in Canada that a third of charges are dropped by prosecutors. Moreover, when cases go to trial, the most common outcome is an acquittal.\footnote{Though, \citet{puddister2021police} a multitude of reasons why these trials often fail to secure a conviction far less than convictions for the general public.} 

\paragraph{Elected officials.} Across the political spectrum, elected officials are generally wont to publicly criticize police, in both the US \citep[page 321,][]{karakatsanis2025copaganda} and Canada \citep[Chapter 2,][]{cole2022skin}.  While ostensibly nonpartisan, police unions in Canada have made overt political endorsements in both countries \citep[Chapter 15,][]{karakatsanis2025copaganda}.\footnote{
Multiple police associations endorsed the Conservative Party of Ontario in 2025: \tiny\url{https://www.cbc.ca/news/canada/toronto/public-safety-courts-jails-bail-police-ontario-election-1.7464093}.} When certain deaths acquire substantial media attention, obliging elected officials to comment, their statements tend to be non-committal. Consider one such statement by Ottawa mayor in 2016 in relation to the death of Abdirahman Abdi, originally excerpted by   \citet[pp. 45,][]{cole2022skin}: 
\begin{quote}
\textit{``Many people have said... `he was murdered!' or `The police killed him!' We just don't know. Unless you were a witness, it's hard to speculate.''}
\end{quote}
This again aligns with prior findings by \citet{sutton_thin_2021} that state official statements are defensive, neutral, but rarely critical of the punishment bureaucracy. 

\subsection{Civilian Sources} \label{sec:civillian-advocates}
The previous section could be understood as entities that tend to be defensive of the criminal punishment bureaucracy. Here, we describe entities that are far more likely to be critical of it, who are sometimes quoted in news reports. These critiques manifest in both direct and indirect ways. The most obviously direct way is to question and undermine excessive force applied by the police. But the other way is to recognize the ``complex personhood'' \citep{white2021whose} of the victims. 

\paragraph{Community members.} As \citeauthor{white2021whose} explains, this is primarily achieved by journalistic labor in identifying and interviewing family, friends, and community peers, higlighting their social embeddedness prior to their deaths. This is in contrast to Canadian police press releases, which do not contain such information \citep{walby_examining_2022}.

Indeed, \citet{sutton_thin_2021} finds that Canadian journalists often interview friends, family, and community peers, with such accounts manifesting in $44\%$ of their sample.\footnote{Though it should be noted that this is significantly less than the number of articles that contain an account from a police officer, not to mention all of the other state bureaucrats listed in the section above.} \citet[][pp. 141-144]{sutton_thin_2021} find that a majority of the accounts from community members are in recognizing their complex personhood -- hobbies, occupations, positive relationships. 
They also note that there are instances wherein community members allude to the victim's past substance abuse or criminal record, though this frame is far less common.




\paragraph{Family lawyers.} \citet{sutton_thin_2021} measures mentions of lawyers to be in 33\% of cases. However, they do not comment on whether these are defence lawyers (defending the police) or family attorneys. Upon inspection of our dataset, introduced in  detail in Section~\ref{subsec:large-data-collection}, we find that a large fraction of these are in fact lawyers hired by the aggrieved family.  Civil-rights lawyers are often hired to represent families in civil litigation, and, more rarely in criminal trials \citep{puddister2021police}. As they are representing the aggrieved family and community, they are predictably critical of the police:
\begin{quote}
``\textit{Then there's the question of racism. I mean, these members [are] seen laughing at him, and someone's got to look at that. There's a real problem in the RCMP.}'' \citep{cbc2010yukonjail}
\end{quote}

\paragraph{Advocacy groups.} Within that dataset, we also find a non-negligible number of advocacy organizations. To name a few, these comprise First Nations organizations (e.g., \textit{Southern Chiefs Organizations}), LGBTQ advocacy groups (e.g., The 519), and homelessness advocacy groups (e.g., \textit{The Old Brewery Mission}):   
\begin{quote}
``\textit{It's the third person suffering from mental health issues in the past three years who has been shot by police.}'' \citep{kelly2014magloire}
\end{quote}
That there is a presence of these organizations makes sense, as they often work with the same historically marginalized groups who are at significantly increased risks of violent interactions with the police \citep{maynard2025policing}. 

\paragraph{Eyewitnesses}
Eyewitness testimony is also not uncommon in these news reports, with \citet{sutton_thin_2021} finding that they manifest in around 12\% of articles in their sample. \citeauthor{sutton_thin_2021} notes that they are among the groups who often raise questions about excessive use of force: 

\begin{quote}
``\textit{Witnesses have said he wasn't armed, but had a cellphone in his hand.''} \citep{kusch2013inquest}
\end{quote}
Though, it is not unprecedented for witnesses to legitimate police force; see \citet[][pp. 48]{cole2022skin} for an example.

\section{\model{} Model} \label{sec:model}
Here we describe our method to calculate the volume of points of view from state bureaucrats  and civilians for reporting on police-involved deaths. Concretely, for an article with $N$ paragraphs $P_1,\dots,P_N$, we obtain labels $y_1,\dots,y_n$, where $$y_i\in\{\text{\paligned},\text{\critical},\text{\context}\}$$. The \paligned{} label reflects that the passage describes the point of view  police official or some other entity or organization in the punishment bureaucracy (\paligned; Section~\ref{sec:punishment-bureaucrats}). The \critical{} label typically applies to passages that articulate the points of view of entities described in Section~\ref{sec:civillian-advocates} (community members, family lawyers, advocacy organizations) who are much more likely to be critical of police agencies and sympathetic to the deceased. The \context{} label applies when the passage is not describing any entity's point of view at all; but rather providing contextual details about the incident, for example, when and where the incident took place.\footnote{Choosing between the \paligned{} or \critical{} label for a given paragraph $P_i$ is largely conditional on the entity providing the perspective, in that if a perspective is from one of the entities in Sec.~\ref{sec:punishment-bureaucrats}, then we apply \paligned{}, and \critical{} if it is one of the entities in Sec.~\ref{sec:civillian-advocates}.}

\paragraph{Datasets.} This detailed parsing is infeasible to accomplish at scale, so we develop and apply the \model{} to accomplish this. We first collect a  dataset with $N=100$ articles, where we annotate  (\S\ref{sec:model:data}) every paragraph in each article with the point of view it represents. For notational simplicity, we assume that each article has $n$ paragraphs; then we have $\mathcal{S} =\{(P_{i}, y_i)\}^{N\cdot n}_{i=1}$. We use this dataset to develop the \model{} model (Section~\ref{sec:model:data}). Then, we measure the volume of perspectives from each group at a much larger scale, with a dataset containing $M=4000$ articles, using labels predicted by \model{}: $\mathcal{U}=\{(P_i, \hat{y_i})\}^{M\cdot n}_{i=1}$ (Section~\ref{sec:large-scale-analysis}).



\paragraph{Contrasting reference with point of view.}
In this work, we distinguish between reference and perspective. For example, the following would be included as a perspective, as here, the \textcolor{violet}{relevant entity} they are expressing: ``\textit{The involved officer did nothing wrong in the incident, the \textcolor{violet}{\textbf{SIU}} concluded...}''. However, the following would  not be included as a perspective from the SIU, as it is context on the standard operating procedure in deadly force incidents, rather than a point of view provided by the SIU: ``The \textcolor{violet}{\textbf{SIU}} investigates all cases where police action results in death.'' See \citet{wiebe_tracking_1994} for discussion what constitutes a perspective or point of view.

\subsection{\model{} Development Dataset}
\label{sec:model:data}

We work with the \href{https://newsinteractives.cbc.ca/longform-custom/deadly-force/}{CBC Deadly Force} database, which documents encounters from 2000-2017. We use this database to search for additional reporting from different outlets on the same incident. In addition, we collected articles for incidents that transpired over the past 10 months (from September 2024 to June 2025) that we learned about through news alerts or social media. This resulted in $N=100$ articles, most from the CBC ($29$), followed by Global News ($10$), CTV ($9$), and a long tail reports from smaller local outlets (for example, \textit{Ottawa Citizen}, \textit{Burnaby now}).


\paragraph{Annotation}
One co-author independently coded these articles, obtaining the articles paragraph-label ($P_i$, $y_i$) pairs.  First, they identified any entity $\epsilon$ mentioned in the article and assigned them as an  \paligned{} or \critical{} label following the definitions in Sec.~\ref{sec:punishment-bureaucrats} and Sec.~\ref{sec:civillian-advocates}. They then identified all passages where the entity $\epsilon$ expresses a perspective: $P_{\epsilon(1)},\dots P_{\epsilon(k)}$, for $\epsilon(i)\in \{1,\dots,n\}$. Any paragraphs where there is no entity expressing a perspective is assigned the \context{} label. This procedure results in the \model{} development dataset denoted by $\mathcal{S}$, containing $2158$ annotations over the $100$ articles. 

\paragraph{Agreement.} Another coauthor then randomly selected $15$ of these articles to code independently. We thus obtained $y_{i}^{1}$ and $y_{i}^{2}$ for these articles, for $i\in[n]$. Over the articles, we obtained 311 paired labels which yielded an agreement of $\kappa=79.4$, indicating substantial agreement. This indicates that $\mathcal{S}$ is a reliable dataset for training \model{}. 


\subsection{Training \model{}} 
\label{sec:model:model}

\paragraph{Modeling approach.}
While it is possible to simply finetune a language model on annotated passages $(P_i, y_i)\in\mathcal{S}$, it is difficult to attribute the point of view expressed in a passage without context from prior passages. Consider Figure~\ref{fig:globe-and-mail-parse}, where the police spokesperson introduced with their full title, clearly indicating that the passage should be assigned the \paligned{} label. Later passages use only their last name, making it difficult to distinguish between \paligned{} and \critical{} without access to prior passages.

We instead train the language model to jointly label a collection of passages $P_{\hat{\epsilon}(i)},\dots,P_{\hat{\epsilon}(k)}$, where $\hat{\epsilon}$ represents a coreference cluster. For example, the coreference cluster for the article in Figure~\ref{fig:globe-and-mail-parse} would be \texttt{RCMP spokeswoman Brigdit Leger}, \texttt{Ms. Leger}, \texttt{Ms. Leger}, and we can obtain the passages containing these references.

\paragraph{Aligning annotated entities and clusters.} To provide supervision to finetune the language model to work with coreference cluster passages, we need to align our manually annotated entities ($\entities$) with the coreference clusters ($\fcorefentities$). We use simple string equality to match our annotated entities ($\entities$) against the coreference clusters produced by \fastcoref{} ($\fcorefentities$). We find that simple string equality captures $71\%$ of relevant paragraphs $P_{\epsilon(1)},\dots,P_{\epsilon(k)}$ for all entities $\epsilon\in\entities$. However, some alignments are missed through exact string matching, for example when the annotated entity is $\epsilon=\text{``\texttt{BEI}''}$, but the coreference cluster only contains instances of the full name of the agency (``\texttt{Bureau des enquêtes indépendantes}''). 

Without repairing these missed alignments, the language model would be prone to labeling relevant passages as mere \context{}, resulting in a high rate of false negatives. We manually repaired these missed alignments, thereby improving recall of the passages with annotated entities in our training set to a more reasonable 91\%. We now have input-output examples of the form $$\left((P_{\hat{\epsilon}(i)},y_{\hat{\epsilon}(i)}),\dots,(P_{\hat{\epsilon}(k)},y_{\hat{\epsilon}(i)})\right)$$ one for each $\hat{\epsilon}\in\fcorefentities{}$,  that we use to finetune the language model. The application of a coreference resolution model followed by predictions on top of the coreference clusters constitutes the \model{} model.


\paragraph{Implementation details.} We first split the $N=100$ articles in $\mathcal{S}$ into a split of $50/25/25$ articles, or $1055/600/503$ annotated paragraphs, for training, validation (hyperparameter tuning), and testing. For replicability and efficiency, we aim to develop a model that can be run with moderate computing resources; we opted for \flantfive{}, a model with a modest (by current standards) 800M parameters. We used a single NVIDIA RTX A6000 GPU to finetune the model. We use \fastcoref{} \citep{otmazgin-etal-2022-f} off-the-shelf for coreference resolution.

\subsection{Assessing the Reliability of \model{}}

\paragraph{Evaluation.} For an article with $N$ paragraphs $P_1,\dots,P_N$, we assess the capacity of the model to predict the point of view reflected in the passage $\hat{y_1},\dots,\hat{y_N}$, comparing against our manual annotations $y_1, \dots, y_N$, where $y_i\in\{\paligned{}, \critical{}, \context{}\}$. We report the F1-score for each possible label.

\paragraph{Baselines.} To contextualize the performance of \model{}, we compare the model with two baselines. We assess GPT-4o's capacity to perform this task. We used the training set for prompt-tuning; see the full prompt in our repository. We also compare against a random classification baseline, where one of the labels is picked uniformly at random.

\begin{table}[t]
\setlength{\tabcolsep}{2pt}
    \begin{tabular}{lcccc}
    \toprule
     \textbf{} & \textbf{\model{}} & \textbf{GPT-4o} & \textbf{Random} & \textbf{$n$}\\
     \midrule
     \paligned{} & $0.76$ & $\mathbf{0.82}$ & $0.39$ & $178$\\  \critical{} & $0.75$ & $\mathbf{0.79}$ & $0.32$ & $115$ \\
     \context{} & $\mathbf{0.77}$ & $0.74$ & $0.40$ & $243$ \\
     \midrule
     overall & $0.76$ & $\mathbf{0.78}$ & $0.38$ & $536$\\
     \bottomrule
    \end{tabular}
    \caption{Classification results for passages across $25$ articles on the test set portion of our manually annotated dataset $\mathcal{S}$. See Section~\ref{sec:model} for more details.}
    \label{tab:f1-classification}
\end{table}

\begin{figure}
    \centering
    \includegraphics[width=0.7\linewidth]{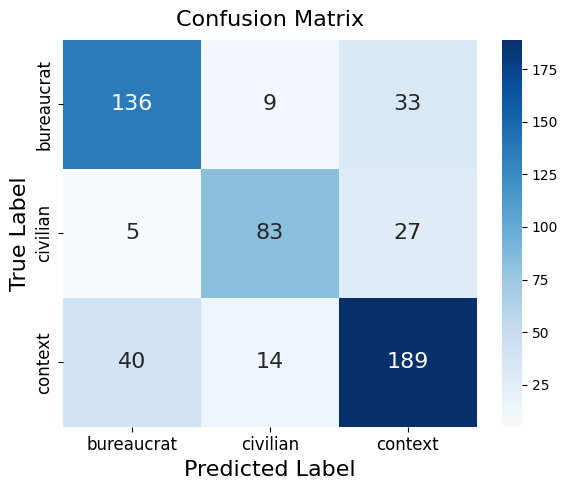}
    \caption{Confusion matrix between the labels.}
    \Description{A 3x3 confusion matrix for a classifier with labels bureaucrat, civilian, and context. The diagonal shows correct predictions: 136 bureaucrat, 83 civilian, and 189 context. The largest off-diagonal errors are context misclassified as bureaucrat (40) and bureaucrat misclassified as context (33).}
    \label{fig:confusion-matrix}
\end{figure}

\paragraph{Performance.} We report the F1-score for each label in Table~\ref{tab:f1-classification}. We find that both \model{} and GPT-4o perform reliably well across all categories. We didn't find a significant difference between the two; a bootstrap test with $B=1000$ samples, comparing the number of times \model{} scored a higher macro-averaged $F_1$ than GPT-4o yielded $p=0.15$ \citep{efron1994introduction}. 
We thus apply \model{} for the remainder of our analyses.

\paragraph{Errors.} We display the confusion matrix in Figure~\ref{fig:confusion-matrix}. We find that the model generally confuses between \context{} and \paligned{}, or \context{} and \critical{}, but rarely \critical{} and \paligned{}, which would be a more severe error. From Figure~\ref{fig:confusion-matrix}, we find that the model does well in capturing the distributional statistics of the labels:
\begin{center}
\begin{tabular}{c|cc}
    \toprule
     \textbf{Label} & \textbf{Annotated} & \textbf{Predicted}  \\
     \midrule
     \paligned{} & 33\% & 33\% \\ 
     \critical{} & 21\% & 20\% \\
     \context{} & 45\% & 46\%\\
     \bottomrule
\end{tabular}
\end{center}

\begin{figure}
    \centering
    \includegraphics[width=\linewidth]{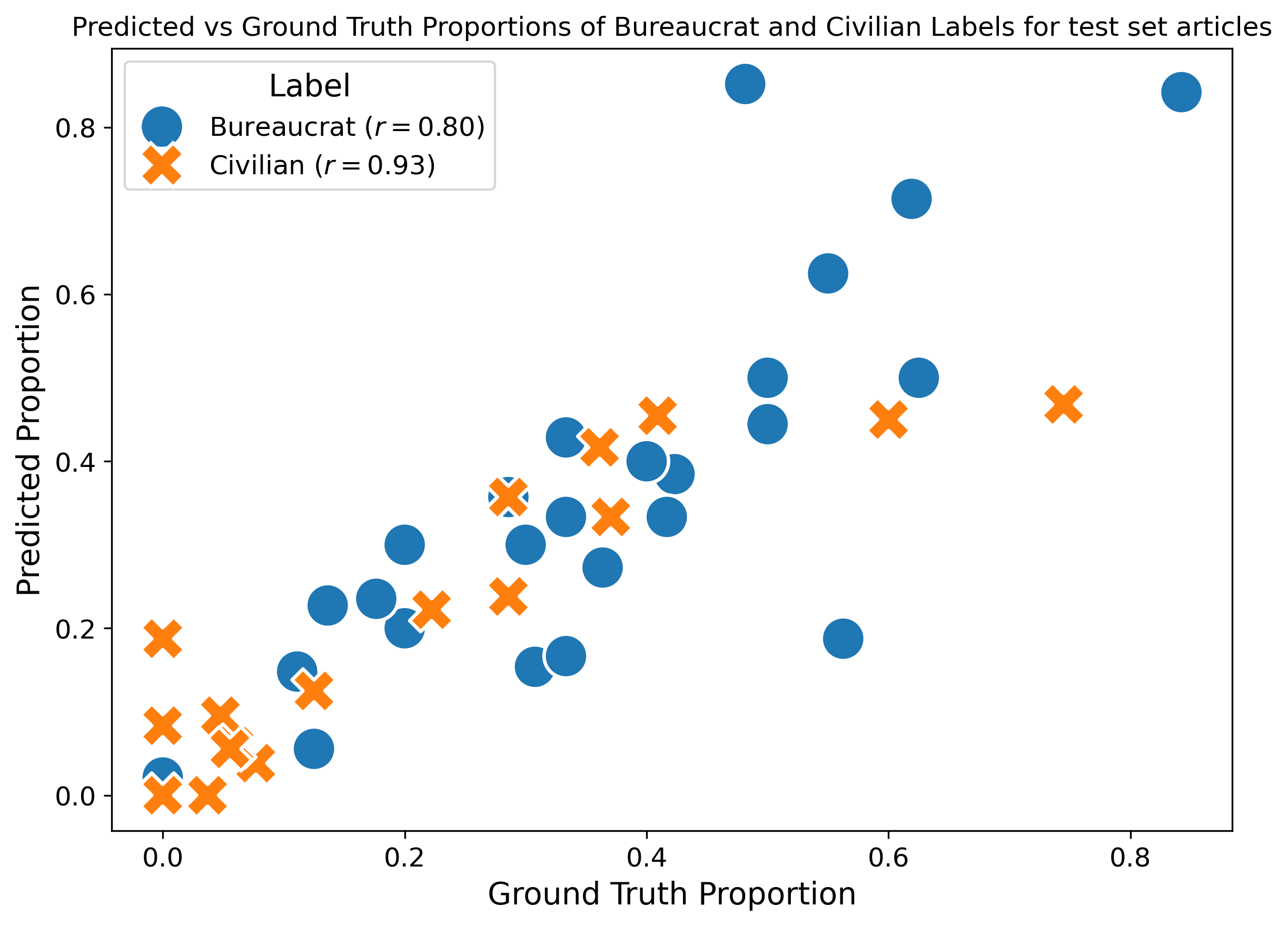}
    \caption{Comparison of predicted vs. ground-truth reference proportion of passages that are labeled as \paligned{} or \critical{} across $25$ articles in the test set.}
    \Description{A scatter plot comparing predicted versus ground truth proportions for two label types across test set articles. Civilian labels (orange crosses, r=0.93) show a strong correlation between predicted and ground truth values. Bureaucrat labels (blue circles, r=0.80) show a moderate-to-strong correlation but with more variance. Both series span the full 0 to 1 range on the ground truth axis.}
    \label{fig:article-level-distribution}
\end{figure}

At the  level of individual articles in the test set, too, we find that \model{} performs well at capturing the distribution of perspectives that are \paligned{} as opposed to \critical{} (Figure~\ref{fig:article-level-distribution}).

\section{\model{} at Scale} \label{sec:large-scale-analysis}
 Having demonstrated the effectiveness of the efficient \model{}, we go onto perform an analysis on a larger-scale corpus, containing $\lvert \mathcal{U}\rvert=4000$ articles. We describe the construction of that dataset next (Section~\ref{subsec:large-data-collection}), followed by analyses on investigating the degree to which state bureaucrats are deferred to in interpreting deadly force incidents with \model{} (Section~\ref{sec:large-scale-analysis}). 

\begin{figure}
    \centering

    \begin{subfigure}{\linewidth}
        \centering
        \includegraphics[width=\linewidth]{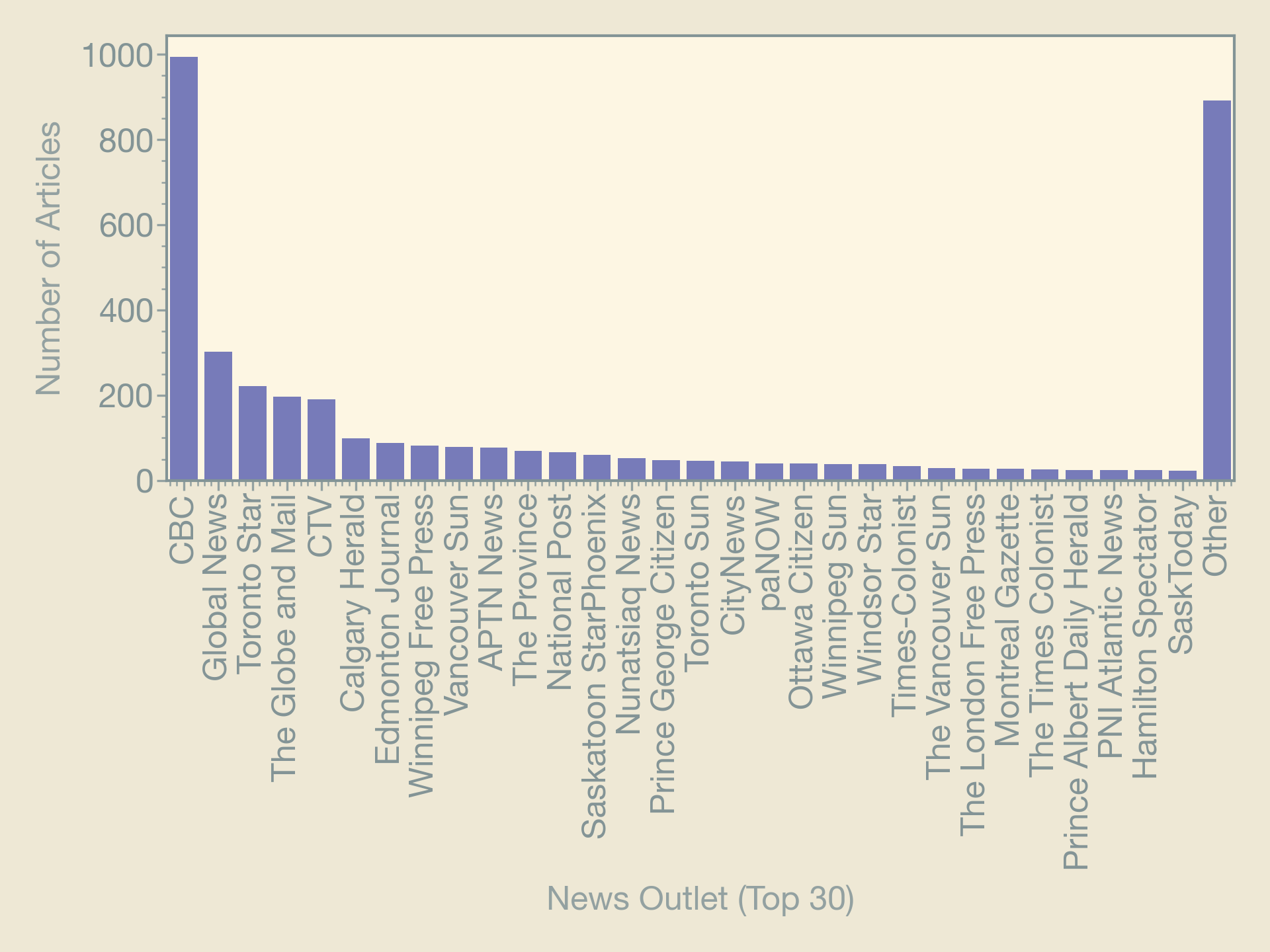}
        \label{fig:outlet-distribution}
    \end{subfigure}

    \vspace{0.5em}

    \begin{subfigure}{\linewidth}
        \centering
        \includegraphics[width=\linewidth]{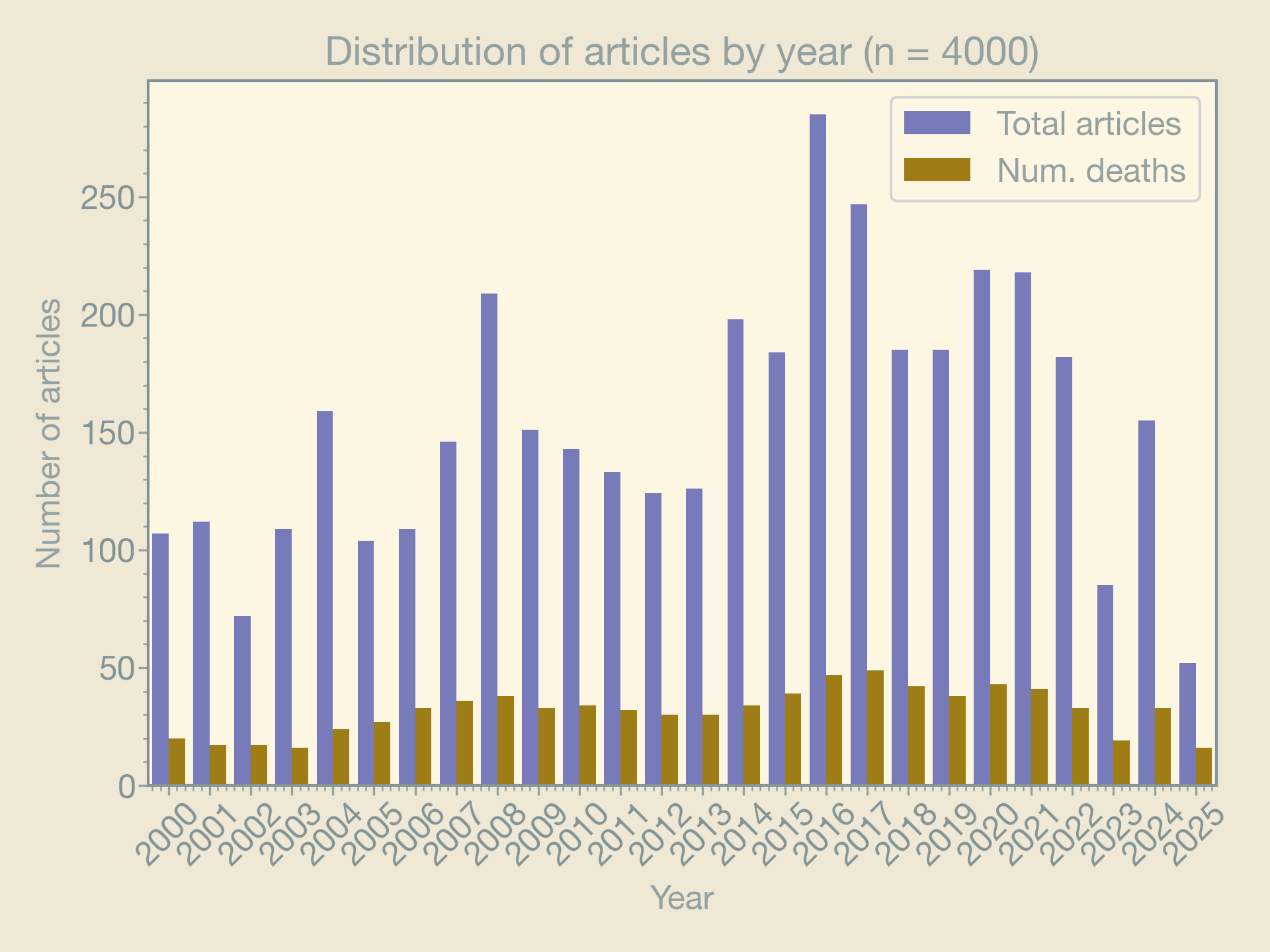}
        \label{fig:year-distribution}
    \end{subfigure}

    \caption{Outlet and year distributions for the large-scale \model{}-annotated dataset $\mathcal{U}$.}
    \label{fig:outlet-year-distributions}
    \Description{Two bar charts stacked on top of each other. First, a bar chart showing the number of articles per news outlet for the top 30 outlets in the dataset. CBC dominates with approximately 1000 articles, followed by Global News (around 300) and Toronto Star and CTV (around 225 each). Articles from all other outlets constitute around 1,000 articles. Second, a grouped bar chart showing the distribution of 4000 articles by year from 2000 to 2025, with a secondary bar showing number of deaths per year. Total articles peak around 2016 at approximately 280, with a secondary cluster of high-volume years from 2018 to 2021. Deaths per year are consistently lower, peaking around 50 in 2016 and 2017, and generally tracking article volume.}

\end{figure}

\subsection{Large-scale data collection} \label{subsec:large-data-collection}
We sourced victim records from three datasets: CBC's Deadly Force, Tracking Injustice \citep{crosby2025tracking}, and the Wikipedia page ``List of killings by law enforcement officers in Canada.''\footnote{\url{https://en.wikipedia.org/wiki/List_of_killings_by_law_enforcement_officers_in_Canada}}. We collected articles and victim records were collected only for incidents occurring in 2000 or later. Only cases with identified victim names were included, leaving us with 811 victim records.
 We then compiled a list of Canadian news outlets was compiled from the \href{https://about.proquest.com/en/products-services/canadian_newsstand/}{ProQuest Canadian Newsstream} title list, 
 restricting the set to publications with internet-accessible content.
 
 For each victim record, we programatically executed queries on search engines using metadata fields including full name, first/last name, age, gender, incident date, province, police service, then checking search results against the aforementioned set of links. We also applied named-entity recognition and regex matching to match articles with victim records. We obtained 4,000 articles across 235 outlets; we show the distribution across outlets and years in Figure~\ref{fig:outlet-year-distributions}. To complete the dataset $\mathcal{U}$,we apply \model{}, obtaining a label in $\{\paligned{},\critical{},\context{}\}$ for every paragraph in every article.

\begin{figure}
    \centering
    \includegraphics[width=\linewidth]{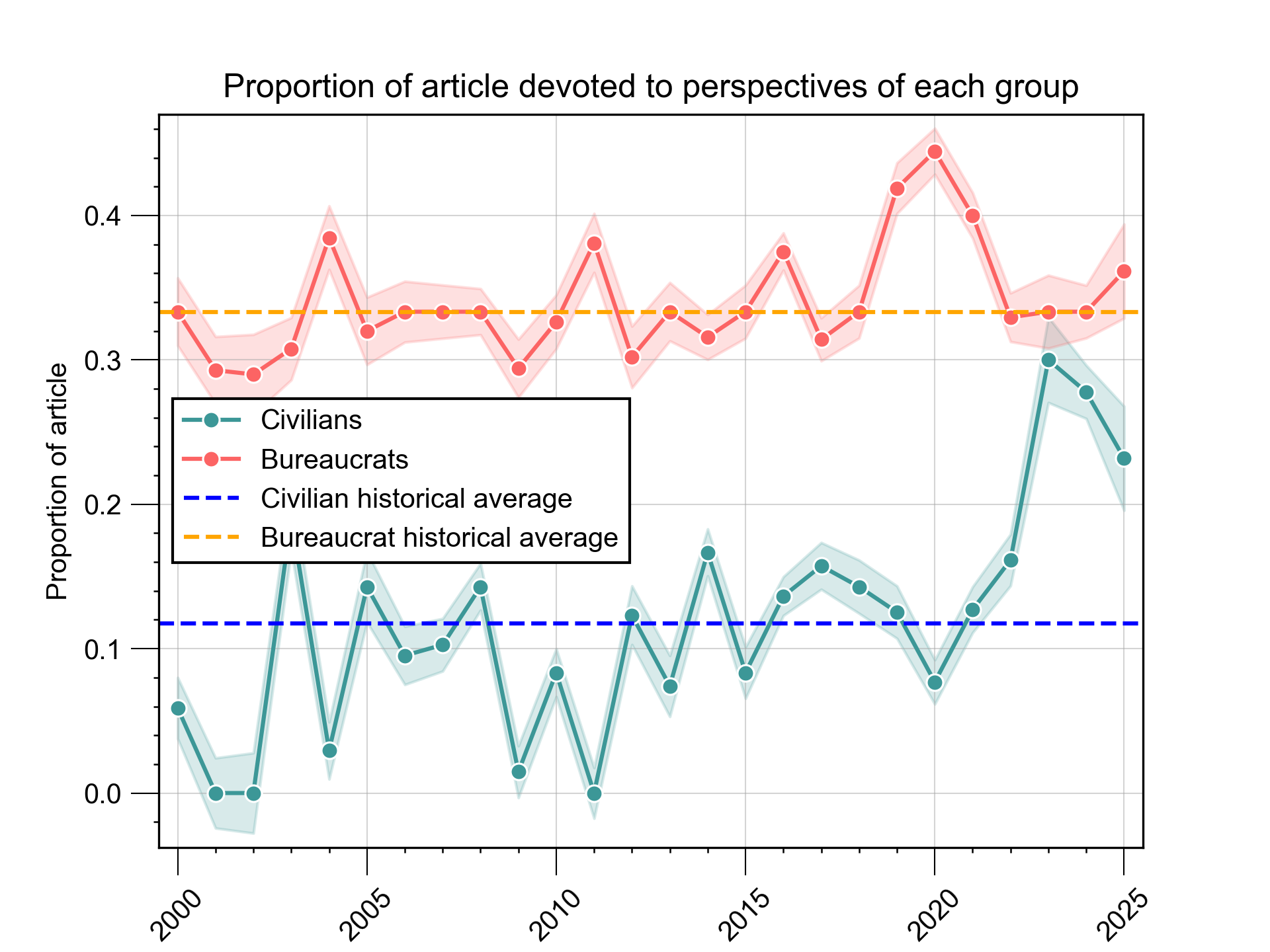}
    \caption{Median proportion of articles devoted to perspectives \paligned{} with state bureaucrats vs. proportion of perspectives for civilians. }
    \Description{A line chart showing the proportion of news articles devoted to the perspectives of civilians and bureaucrats from 2000 to 2025. Bureaucrats consistently receive a higher share, fluctuating around their historical average of 0.33, with a notable peak near 0.45 around 2020. The civilian share is lower and more volatile, hovering around a historical average of 0.12, but rises sharply after 2022 to approach 0.30 before declining again in 2025. Confidence intervals are shown as shaded bands around each line.}

    \label{fig:placeholder}
\end{figure}
\subsection{Quantitative Analysis}
\paragraph{News reporting on deadly force incidents has been highly technocratic historically.} Over the last 25 years, state bureaucrats  have had much greater epistemic authority in narrating deadly force incidents compared to civilian-based accounts. Quantitatively, we find that the median article over the entire time period has $33.3\%$ of its passages reflecting perspectives of state bureaucrats. Comparatively, only $11.8\%$ of articles are devoted to civilian-based accounts. In their exposition of media representation of crime, \citet{chan2014racialization} wrote that ``police…arguably remain the first authorities to speak about, and therefore frame, the issues in most mainstream media accounts of crime.'' Our analysis shows that their explanation holds case even in police-involved deaths, where it is the police themselves who face scrutiny for potential extrajudicial use of force. 

\paragraph{Since 2020 reporting has become more co-productive.} Our results suggests that epistemic authority in news reporting is contingent on political debates and public pressure, with civilian-based accounts having a sharp increase in coverage of their perspectives from 2020, peaking at $30\%$ in 2023. It is hard to ignore that this increased representational volume of civilian-based accounts coincides with increased nationwide scrutiny of systematic police misconduct. In 2020, for example, 51\% of Canadians expressed support for reducing police funding \citep{becken2023defund}.
We also observed the construction of not one but two national databases for tracking deadly force incident \citep{marcoux2018deadly,crosby2025tracking}. Our analysis suggests that news reporting patterns changed, such that increased journalistic labor was invested in soliciting civilian-based accounts.



Overall, our result demonstrates novel evidence that this time period resulted in quantifiable gains to the epistemic authority afforded to civilians. We thus observe a greater plurality of interpretations of these incidents, thus aligning with a more co-productive form of public opinion formation as espoused by political philosophers \citep{dewey_public_2012,ostrom1996crossing} and media theorists \citep{rosen1999journalists}. Whether this increased attention to civilian-based accounts persists beyond this time period remains to be seen; we observe civilian accounts trending downwards towards the historical average again for 2024 and 2025.  

\paragraph{Across-outlet variation.} For brevity, we look at the top-5 outlets in terms of number of articles: CBC, Toronto Star, The Globe and Mail, Global News, and CTV News. We observe limited variation in terms of coverage devoted to state bureaucrats, with CBC having the highest rate ($35\%$), and CTV the lowest ($31\%$). By comparison, coverage devoted to civilians varies substantially, with the Toronto Star at the highest, well above the national median at $23.2\%$. This is followed by the CBC ($16\%$) and the Globe and Mail ($15\%$). CTV and Global News fall well below the national average, with the median article having $2.5\%$ and $0\%$ of its passages devoted to perspectives of civilians, respectively. 

We find that outlets consistently include points of view from state bureaucrats; there is little cross-outlet variation on this point. This is aligned with the fact that state bureaucrats have established relationships with reporters \citep{mcgovern2010cop}, making it easier to ascertain and report on their perspectives. Moreover, bureaucrats have an interest in conveying their perspectives to the media as a public relations risk-mitigation strategy \citep{walby_examining_2022}.

Where we find far larger variation is whether outlets include civillian-based accounts.  
We find that the Toronto Star has a greater degree of coverage devoted to civilian-based accounts. This is aligned with the history of the Star's reporting  interests in scrutinizing and challenging points of view from state bureaucrats; for example, they conducted the largest known investigation of racial bias in police investigations in 2002 \citep{owusu-bempah_black_2014,rankin2012knowntopolice}. 

Meanwhile, CTV and Global News are generally understood to serve a different role in Canadian newsmedia, with more episodic, breaking-news style content. This can be evinced by the fact that their articles tend to be shorter (median length: 13 and 14 paragraphs, respectively), well below the national median of $16$ paragraphs. Meanwhile, the other three outlets are above the national median. Producing shorter content at a faster pace necessarily lends itself to relying on press releases that are prepackaged by Police Media Units \citep{mcgovern2010cop,karakatsanis2025copaganda}, rather than performing the journalistic labor required to get civilian accounts \citep{white2021whose}.

\subsection{Qualitative Analysis}
Our quantitative analysis established that state bureaucrats' points of view receive much more representation than civilians in reporting on police-involved deaths. Here we look at how their points of view differ in terms of content. Since there are tens of thousands of passages for both groups, we summarize major discursive differences by identifying their distinct key terms. 

\begin{figure}
    \centering
    \includegraphics[width=\linewidth]{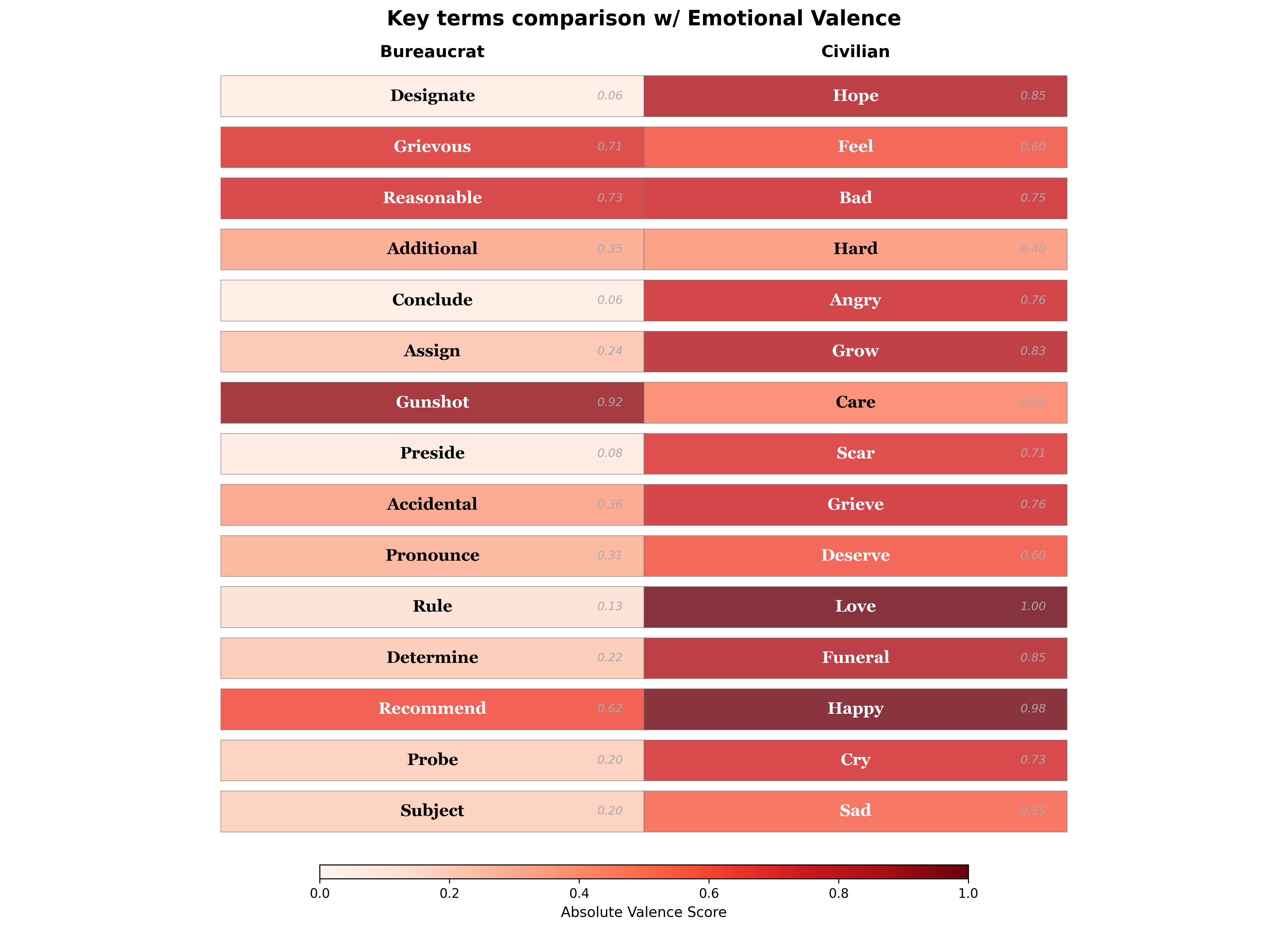}
    \caption{Key terms identified for passages tagged as \paligned{} (left) vs. \critical{} (right). The terms are sorted by their log-odds scores, relative to a Dirichlet prior distribution parameterized by their frequencies in \context{} passages.}
    \Description{A paired bar chart comparing the emotional valence of key terms associated with Bureaucrat and Civilian language, with bars color-coded from light pink (low valence) to dark red (high valence). Bureaucrat terms are predominantly procedural and low-valence (e.g., Designate: 0.06, Conclude: 0.06, Preside: 0.08), with exceptions like Gunshot (0.92) and Grievous (0.71). Civilian terms are consistently higher in emotional valence, including Love (1.00), Happy (0.98), Hope (0.85), and Funeral (0.85), reflecting more emotionally charged language overall.}
    \label{fig:discourse-terms}
\end{figure}

Specifically, we examine the terms that are over-represented in \paligned{} discourse compared to \critical{} discourse (and vice-versa). We compute the distinctiveness of each verb across the two groups of passages, computing their log-odds ratio with a Dirichlet prior, also known colloquially as Fightin' Words; see \citep{jurafsky2024speech,wanner2025does}. We use \context{} passages to parameterize the Dirichlet prior. We list the top-15 terms for both groups in Figure~\ref{fig:discourse-terms}, with each term shaded by it's emotional valence score from the NRC VAD lexicon \citep{mohammad2025nrc}.

\subsubsection{Key discourse terms -- \paligned{} passages}. We find that \paligned{} passages are largely centered on administrative procedure (\textit{designate probe, preside, assign}), reflecting previous findings that Police Media Unit communications following a police shooting tend to be ``minimalist and risk-averse'' \citep{walby_examining_2022}. These terms largely lack emotional valence, as measured by the NRC VAD lexicon \citep{mohammad2025nrc}. Even terms carrying higher emotional valence (\textit{grievous, reasonable, gunshot, recommend}) are strongly related to administrative procedure. For example, \textit{gunshot} typically refers to the victim's cause of death (as determined by a coroner); \textit{reasonable} is used uttered when oversight agencies determine that (in their view) the officer's conduct was warranted. Similarly, \textit{grievous} is a legal designation,\footnote{\url{https://en.wikipedia.org/wiki/Grievous_bodily_harm}} often invoked to argue that the officer believed they were under threat of grievous bodily harm and therefore warranted to use deadly force:

\begin{quote}
\textit{When pressed on the choke hold she was using, Taylor said ``it is a last resort to be used when there is a threat of death of grievous harm.''} \citep{davidson2009man}
\end{quote}

\subsubsection{Key discourse terms -- \critical{} passages} By comparison, the terms that characterize \critical{} discourse carry a much higher degree of emotional valence. Civilians, specifically families, often express their hope for justice, that the administrative oversight procedures result in some sort of penalty for disproportionate conduct: 
\begin{quote}
\textit{``There is always hope that the system will bring justice for a person killed by police, but the reality never lives up to this hope as the system is stacked against us,'' reads a Nov. 16 NTC media release.} \citep{omalley2022police}
\end{quote}

When civilians are quoted, we also tend to get a more vivid and complete picture of the person who was killed, highlighting their complex personhood \citep{white2021whose}, specifically in regards to the social relations they had with others who will be forced to \textit{grieve} them:

\begin{quote}
\textit{For Fagan, the grief of losing her son was overwhelming, and not knowing what caused his death left her unable to eat, sleep or focus. She took a bereavement leave. When she returned to her job she struggled to concentrate. \citep{russell2022chadd}}
\end{quote}
There are also of course expressions of outrage and \textit{anger} towards the jurisdictional police service and the broader punishment bureaucracy:

\begin{quote}
\textit{``Norm Assiff, who is representing Hanna's estate, said Hanna's sister, Susan Bandola, is "angrier and left with more questions than answers" in the wake of the ASIRT report.''} \citep{wakefield2024asirt}
\end{quote}

Our quantitative findings demonstrate these civilian points of view have been historically marginalized in reporting on police-involved deaths.

\section{Conclusion}
Drawing on decades of sociology research in Canada mapping relationships between police and media discourse, we develop a novel computational framework for measuring perspective weighting in Canadian news reporting on police involved deaths, as well as a model that performs well at perspective detection. Applying the model on a large dataset of articles about police-involved deaths from 2000-2025, we find that state bureaucrats' perspectives are overwhelmingly deferred to in Canadian news reporting, at a rate of 3 times on average that of civilian accounts. Nevertheless, we find that the proportion of coverage devoted to civilian-based accounts increased after 2020, suggesting heightened visibility of police brutality incidents may have inclined news organizations and journalists to seek more civilian accounts. We anticipate that \model{} can be used to keep track of the dynamics of epistemic authority over the years, as the rate of deadly force incidents is not expected to decline for the foreseeable future \citep{rutland_canadian_2023}.

\bibliographystyle{ACM-Reference-Format}
\bibliography{custom}


\end{document}